\def\year{2023}\relax
\documentclass[letterpaper]{article} 
\usepackage{aaai21}  
\usepackage{times}  
\usepackage{helvet} 
\usepackage{courier}  
\usepackage[hyphens]{url}  
\usepackage{graphicx} 
\usepackage{graphicx}  
\usepackage{natbib}  
\usepackage{caption} 
\usepackage{amsmath}
\usepackage{mathtools,amssymb}
\nocopyright

\usepackage{newfloat}
\usepackage{listings}
\DeclareCaptionStyle{ruled}{labelfont=normalfont,labelsep=colon,strut=off} 
\usepackage[utf8]{inputenc} 
\usepackage[T1]{fontenc}    
\usepackage{hyperref}       
\usepackage{url}            
\usepackage{booktabs}       
\usepackage{amsfonts}       
\usepackage{nicefrac}       
\usepackage{microtype}      
\usepackage{xcolor}         
\usepackage{times}
\usepackage{soul}
\usepackage{tabularx}
\usepackage{bbm}
\usepackage{dsfont}
\usepackage{wrapfig}
\usepackage{graphicx}
\usepackage{multirow}
\usepackage{multicol}
\usepackage{makecell}
\usepackage{amsmath,amssymb,amsthm,amsfonts,mathrsfs,bm,multirow}
\usepackage{subfigure}
\usepackage{pifont}
\usepackage{float}
\usepackage[normalem]{ulem}

\makeatletter
\newcommand*{\rom}[1]{\expandafter\@slowromancap\romannumeral #1@}
\makeatother
\usepackage{enumitem}
\usepackage{comment}

\usepackage{algorithmic}
\usepackage{diagbox}
\usepackage{colortbl}

\definecolor{light-gray}{gray}{0.92}

\newcommand{\squishlist}{
    \begin{list}{$\bullet$}
        { \setlength{\itemsep}{0pt}      \setlength{\parsep}{0pt}
            \setlength{\topsep}{0.5pt}       \setlength{\partopsep}{0pt}
            \setlength{\listparindent}{-2pt}
            \setlength{\itemindent}{-5pt}
            \setlength{\leftmargin}{0.5em} \setlength{\labelwidth}{0em}
            \setlength{\labelsep}{0.2em} } }
    
\newcommand{\squishend}{
\end{list}  }

\usepackage[capitalize,noabbrev]{cleveref}
\usepackage[textsize=tiny]{todonotes}

\title{Peeling the Onion: Hierarchical Reduction of Data Redundancy for Efficient Vision Transformer Training}
\author {
    Zhenglun Kong \thanks{These authors contributed equally} \textsuperscript{\rm 1},
    Haoyu Ma  \footnotemark[1] \textsuperscript{\rm 2},
    Geng Yuan \footnotemark[1] \textsuperscript{\rm 1},
    Mengshu Sun \textsuperscript{\rm 1},
    Yanyue Xie \textsuperscript{\rm 1},
    Peiyan Dong \textsuperscript{\rm 1}, \\
    Xin Meng \textsuperscript{\rm 3}, 
    Xuan Shen  \textsuperscript{\rm 1},
    Hao Tang  \textsuperscript{\rm 4},
    Minghai Qin \textsuperscript{\rm 5} 
    Tianlong Chen  \textsuperscript{\rm 6} 
    Xiaolong Ma  \textsuperscript{\rm 7}  \\
    Xiaohui Xie  \textsuperscript{\rm 2}
    Zhangyang Wang  \textsuperscript{\rm 6}
    Yanzhi Wang  \textsuperscript{\rm 1} \\
}
\affiliations {
    \textsuperscript{\rm 1} Northeastern University, 
    \textsuperscript{\rm 2} University of California, Irvine, 
    \textsuperscript{\rm 3} Peking university, 
    \textsuperscript{\rm 4} ETH Zurich, 
    \textsuperscript{\rm 5} Western Digital Research,
    \textsuperscript{\rm 6} Unversity of Texas at Austin,
    \textsuperscript{\rm 7} Clemson University
    
    \textsuperscript{\rm 1} \{kong.zhe,yuan.geng\}@northeastern.edu, 
    \textsuperscript{\rm 2} haoyum3@uci.edu,
}

\begin{document}

\maketitle

\begin{abstract}
Vision transformers (ViTs) have recently obtained success in many applications, but their intensive computation and heavy memory usage at both training and inference time limit their generalization. Previous compression algorithms usually start from the pre-trained dense models and only focus on efficient inference, while time-consuming training is still unavoidable. 
In contrast, this paper points out that the million-scale training data is redundant, which is the fundamental reason for the tedious training. To address the issue, this paper aims to introduce sparsity into data and proposes an end-to-end efficient training framework from three sparse perspectives, dubbed \textit{Tri-Level E-ViT}. Specifically,  we leverage a hierarchical data redundancy reduction scheme, by exploring the sparsity under three levels: number of training examples in the dataset, number of patches (tokens) in each example, and number of connections between tokens that lie in attention weights. 
With extensive experiments, we demonstrate that our proposed technique can noticeably accelerate training for various ViT architectures while maintaining accuracy.  
Remarkably, under certain ratios, we are able to improve the ViT accuracy rather than compromising it. For example, we can achieve 15.2\% speedup with 72.6\% (+0.4) Top-1 accuracy on Deit-T, and 15.7\% speedup with 79.9\% (+0.1) Top-1 accuracy on Deit-S. This proves the existence of data redundancy in ViT. Our code is released
at \url{https://github.com/ZLKong/Tri-Level-ViT}
\end{abstract}
\section{Introduction}




After the convolutional neural networks (CNNs) dominated the computer vision field for more than a decade, the recent vision transformer (ViT) \cite{dosovitskiy2020image} has ushered in a new era in the field of vision \cite{hudson2021ganformer,chen2021pix2seq,kim2021hotr,Deng_2021_ICCV,xue2021transfer,zhao2021point,Guo_2021,srinivas2021bottleneck}. 
Existing ViT and variants, despite the impressive empirical performance, suffer in general from large computation effort and heavy run-time memory usages \cite{liang2021swinir,chen2021crossvit,carion2020end,dai2021up,amini2021t6d,misra2021-3detr,chen2021transformer,el2021training, yang2020learning,chen2021pre,lu2021efficient}.
To reduce the computational and memory intensity of the ViT models, many compression algorithms have been proposed \cite{ryoo2021tokenlearner, pan2021iared2, liang2022evit,yu2022unified,kong2021spvit}.
These methods mainly target efficient inference, and they are either conducted during the fine-tuning phase with a pre-trained model or require the full over parameterized model to be stored and updated during training \cite{rao2021dynamicvit, kong2021spvit}.

\begin{figure}[t]
\begin{center}
\centerline{\includegraphics[width=0.9 \columnwidth]{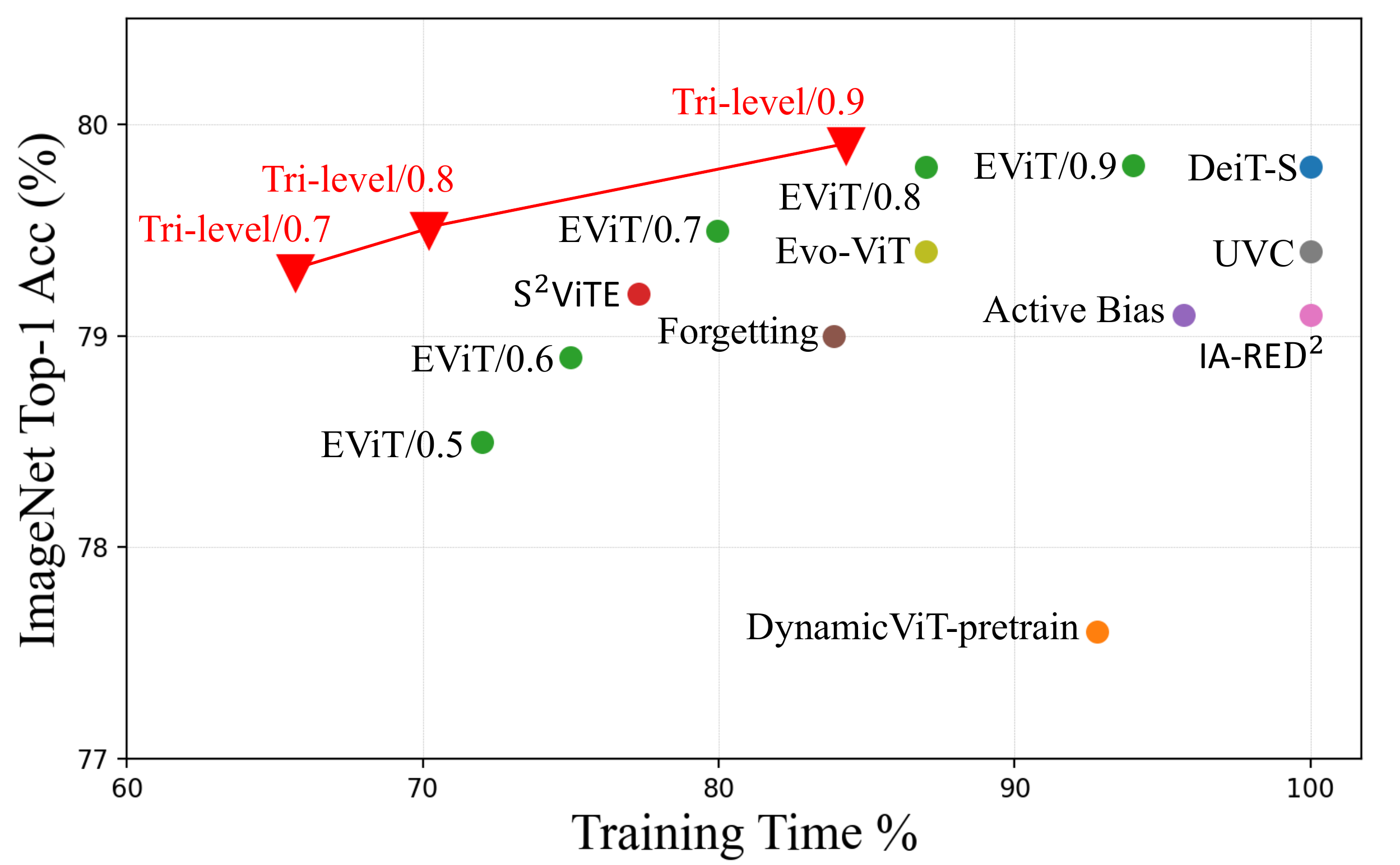}}
\caption{ Comparison of different models with various accuracy-training efficiency trade-off. The proposed Tri-level method achieves a better trade-off than the other methods.}
\label{fig:plot_time_acc}
\end{center}
\vskip -0.2in
\end{figure}

The self-attention mechanism in ViTs abandons the inductive bias that is inherent in CNNs, making the training of ViTs extremely data-hungry.
Specifically, the early ViT work \cite{dosovitskiy2020image} requires to be trained on a much larger dataset (i.e., JFT-300M with $300$ million images) to claim its superiority compared to CNNs that are directly trained using the ImageNet dataset. 
This issue is partially mitigated by the following works, which incorporate knowledge distillation techniques to transfer inductive bias from a convolutional teacher model during training \cite{Touvron2021TrainingDI} or introduce the design of CNNs into transformer blocks \cite{liu2021swin, heo2021pit}.
However, the time-consuming training process still remains computational and memory intensive, and even more costly than before, becoming a huge obstacle to a vast number of ViT-based research and applications.
Therefore, an efficient training paradigm for ViT models is in demand more than ever.
Motivated by such needs, we raise a question: \textit{Can we incorporate data sparsity into the training process to amortize the high training costs caused by the data-hungry nature of the ViTs?}

Due to the quadratic image patch-based computation complexity in the self-attention mechanism of ViT models \cite{dosovitskiy2020image}, 
it is desirable and reasonable to explore the sparsity by leveraging the redundancy that inherent in natural image data. 
Towards this end, we propose an efficient ViT training framework, named Tri-Level E-ViT, which hierarchically eliminates the data redundancy in different levels of the ViT training, \underline{\textit{just like peeling the onion.}}
Specifically, to reduce the training costs and accelerate the training process, we explore a tri-level data sparsity, which includes \textit{example-level sparsity}, \textit{token-level sparsity}, and \textit{attention-level sparsity}. 

Prior works have only focused on removing redundant patches (tokens) instead of removing the whole image (training example) \cite{rao2021dynamicvit,liang2022evit,kong2021spvit,xu2021evo}. 
This is due to the fact that the tendency to train a ViT with a large amount of data has become “deeply rooted” and led people to overlook the existence of redundancy in the example level.
Motivated by this, we design a ViT-specific online example filtering method, to assess the importance of each example and to remove less important ones during training. More specifically, we use both the output classification logit and the variance of the attention map in the \texttt{[CLS]} token \cite{dosovitskiy2020image} to evaluate the importance of each example. We train the model with a subset of the full training examples (e.g., 80\%) and employ a remove-and-restore algorithm to continuously update the training subset, ensuring both the example-level sparsity and optimization throughout the entire ViT training process.

The ViT decomposes an image into several non-overlapping patches. 
The patch-based representation \cite{trockman2022patches} preserves locality and detailed information of an image. Thus, the patch can be considered a type of fine-grained data. 
The ViT conducts the dense self-attention among all patches, which leads to quadratic complexity with respect to the number of patches. 
Recent works suggest that not all patches are informative \cite{rao2021dynamicvit} and not all attention connections among patches are necessary \cite{liu2021swin,heo2021pit}. 
Thus, we further explore two levels of data sparsity from the perspective of the patch: One is token-level sparsity, which aims to reduce the number of patches (tokens), and the other is attention-level sparsity, which aims to remove redundant attention connections. 
We propose the Token\&Attention selector (TA-selector) to simultaneously achieve both levels of sparsity during training. In detail, the TA-selector directly selects informative tokens and critical attention connections based on the attention maps of early layers. 
It is a data-dependent scheme and does not introduce any new weights. Thus, TA-selector is flexible to be installed into many ViT variants.


By incorporating the tri-level data sparsity, our framework significantly reduces the training costs and accelerates the training process, while maintaining the original accuracy. 
To the best of our knowledge, we are the first to explore all levels of data sparsity for efficient ViT training. As shown in Figure \ref{fig:plot_time_acc}, our method achieves the best accuracy-efficiency trade-offs compared to existing sparsity methods.
Most importantly, our Tri-Level E-ViT framework is efficient and practical since we do not require to introduce extra networks during training, such as using a CNN teacher for distillation \cite{Touvron2021TrainingDI} or a predictor network for token selection \cite{rao2021dynamicvit}.
In addition to the training acceleration, the data-sparse ViT model obtained by our framework can also be directly used for efficient inference without further fine-tuning. To verify the robustness, we evaluate the training acceleration of our framework on general-purpose GPU and FPGA devices.



Our contributions are summarized as follows:

\squishlist{}
 \item We propose an efficient framework that employs tri-level data sparsity to eliminate data redundancy in the ViT training process.

 \item We propose an attention-aware online example filtering method specifically for ViT to generate example-level sparsity. We use a remove-and-restore approach to ensure data efficiency throughout the entire training process while optimizing the reduced dataset.
 
 \item We propose a joint attention-based token\&attention pruning strategy to simultaneously achieve both token and attention connection sparsity during training. Our method does not require a complex token selector module or additional training loss or hyper-parameters.



\item We conduct extensive experiments on ImageNet with Deit-T and Deit-S, and demonstrate that our method can save up to 35.6\% training time with comparable accuracy.
We also evaluate the training acceleration on both GPU and FPGA devices.

\squishend{}

\section{Related Work}
\label{relate}

\begin{figure*}[t]
\vskip 0.2in
\begin{center}
\centerline{\includegraphics[width=0.8\textwidth]{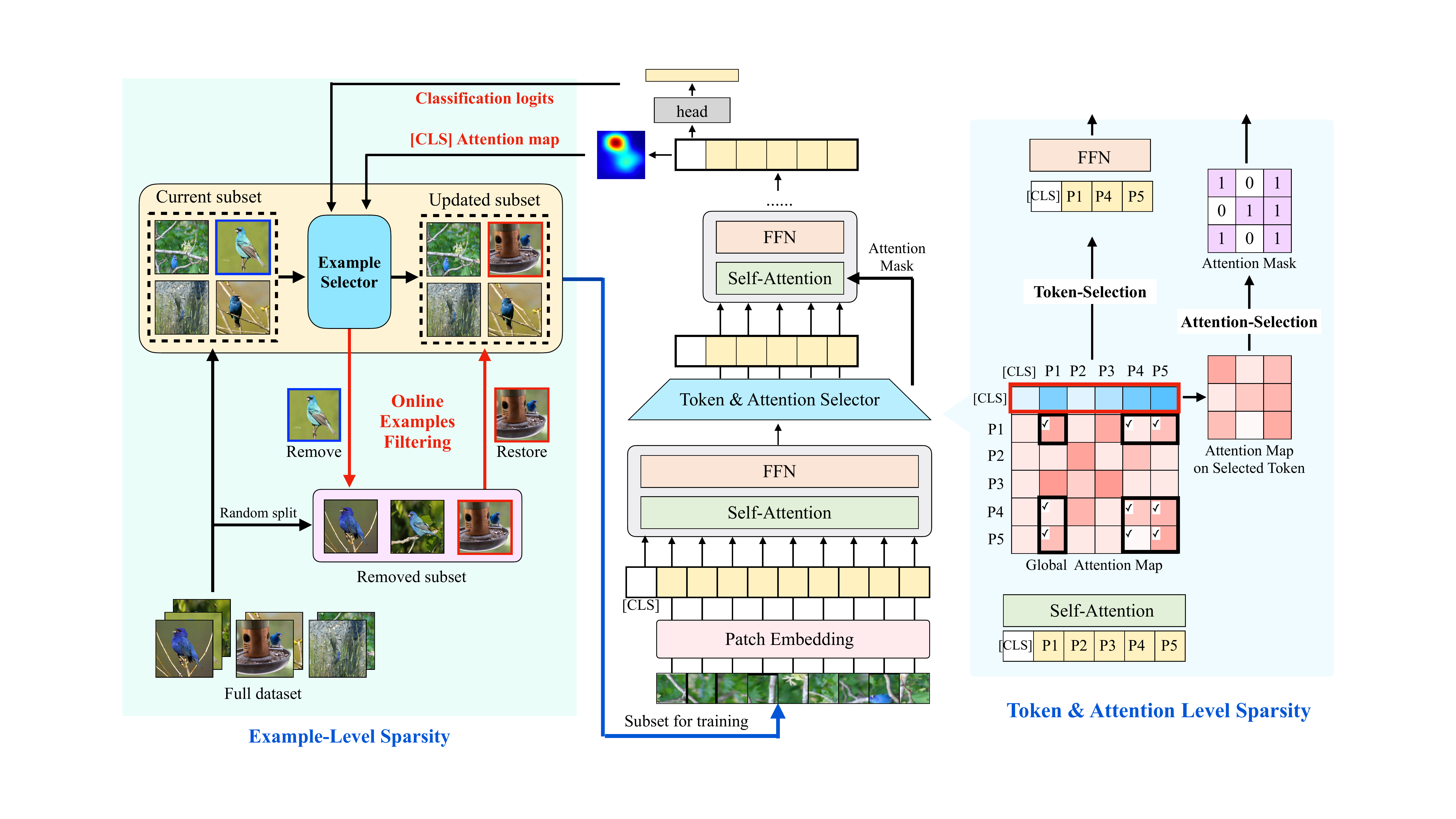}}
\caption{ \small{The overall procedure of our proposed Tri-Level E-ViT framework. We explore the training sparsity under three levels: \textbf{Left}: \textit{Example-Level Sparsity}. Randomly remove an amount of examples prior to training. During the training process, the example selector updates the training subset by removing the most unforgettable examples from the training data and restore the same amount of data from the portion removed before training. \textbf{Right}: \textit{Token \& Attention level sparsity}. Evaluate token importance by the \texttt{[CLS]} token and prune the less informative tokens. We further use the attention vector of each remaining token to decide critical attention connections.} }
\label{ovr-frame}
\end{center}
\vskip -0.4in
\end{figure*}

\noindent\textbf{Vision Transformers.}
ViT \cite{dosovitskiy2020image} is a pioneering work that uses transformer-only structure to solve various vision tasks. Compared to traditional CNN structures, ViT allows all the positions in an image to interact through transformer blocks whereas CNNs operated on a fixed sized window with restricted spatial interactions, which can have trouble capturing relations at the pixel level in both spatial and time domains. Since then, many variants have been proposed. 
For example, DeiT~\cite{Touvron2021TrainingDI}, T2T-ViT~\cite{Yuan_2021_ICCV}, and Mixer \cite{tolstikhin2021mlp} tackle the data-inefficiency problem in ViT by training only with ImageNet. PiT \cite{heo2021pit} replaces the uniform structure of transformer with depth-wise convolution pooling layer to reduce spacial dimension and increase channel dimension.  SViTE \cite{chen2021chasing} alleviates training memory bottleneck and improve inference efficiency by co-exploring input token and attention head sparsity. 
Recent works have also attempted to design various self-supervised learning schemes for ViTs to address the data-hungry issue. BEiT \cite{bao2021beit} and PeCo \cite{dong2021peco} pre-train ViT by masked image modeling with an image tokenizer based on dVAE that vectorizes images into discrete visual tokens. MAE \cite{he2021masked} eliminates the dVAE pre-training process by reconstructing pixels, in contrast to predicting tokens. 

\noindent\textbf{Data Redundancy Reduction.}
There are many attempts to explore redundancy in data. For example level, 
\cite{katharopoulos2018not} proposed an importance sampling scheme based on an upper bound to the gradient norm, which is added to the loss function. \cite{chang2017active} emphasizes high variance samples by reweighting all the examples. However, the additional computation overhead makes them less efficient for training. \cite{toneva2018empirical} used the number of forgotten counts of training examples as a metric to reflect the complexity (importance) of the examples. However, it requires a two-phase training approach that the partial training examples are only removed in the second phase, which limits the overall acceleration. Moreover, all the existing methods are applied only to CNN, making them less suitable for ViT, because their proposed criteria do not consider the ViT architecture for evaluation. 
For Patch level (token/attention), DynamicViT \cite{rao2021dynamicvit} removes redundant tokens by estimating their importance score with a MLP \cite{vaswani2017attention} based prediction module.  Evo-ViT \cite{xu2021evo} develops a slow-fast token evolution method to preserve more image information during pruning.  DAT \cite{https://doi.org/10.48550/arxiv.2201.00520} selects important key and value pairs in self-attention with a deformable self-attention module.  However, all of these methods require additional selectors to guide the removal of redundant data, which is not efficient for training.  EViT \cite{liang2022evit} provides a token reorganization method that reduces and reorganizes the tokens. However, it does not take into account example level or attention level redundancy.

\section{Methodology}
\label{method}

\subsection{Overall Framework}
To reduce the redundant computing caused by the training data, we propose the Tri-Level E-ViT. The overall framework is shown in \cref{ovr-frame}. Specifically, we introduce the sparsity relevant to the data  from three levels: The training example level, the token level, and the attention connection level. All of these sparse methods are combined together during the training process of ViTs.

\subsection{Sparsity in Training Examples}


Prior works \cite{toneva2018empirical, NEURIPS2021_ac56f8fe} show that removing some less informative and easy examples from the training dataset will not lesion the performance of the trained CNN model. The amount of information (complexity) of the examples can be identified by recording the number of forgotten events of examples during CNN training. Examples with lower forgetting counts are generally considered easy to be learned and less informative than others, and can therefore be removed in order. More discussion are shown in the Appendix.
We intend to incorporate this idea into our example-level sparse ViT training.
We face two challenges: 
\textit{1) How to design such an example evaluation method for ViT, considering its unique model characteristics?  
2) How to identify and remove easy examples from the dataset during training to reduce the training time, without prior knowledge of training examples?}

\subsubsection{Attention Aware Example Selector}
We learn the conditional probability distribution given a dataset $D = (x_i; y_i)_i$ of observation/label pairs. For example $x_i$, the predicted label (classification logits) obtained after $t$ steps of AdamW is denoted as $\hat{y}_i^t= \arg\max_k p(y_{ik}|x_i;\theta^t)$. Let $corr_i^t = \mathbb{B}_{\hat{y}_i^t = \hat{y}_i}$ be a binary variable indicating whether the example is correctly classified at time step. When example $i$ gets misclassified at step $t {+} 1$ after having been correctly classified at step $t$,  example $i$ undergoes a forgetting event. The binary variable $corr_i^t$ decreases between two consecutive updates: $corr_i^t > corr_i^{t+1}$. On the other hand, a learning
event occurs if $corr_i^t < corr_i^{t+1}$.

After counting the number of forgetting events for each example, we sort the examples based on the number of forgetting events they undergo. However, there exist examples with the same number of forgetting events. Even if the number is the same, the complexity of the images is still different. Therefore, in addition to the classification logits, we employ the unique self-attention mechanism in the transformer to better measure the complexity of images. We calculate the variance of the attention map of the \texttt{[CLS]} token \cite{dosovitskiy2020image} to obtain the similarity of the image patches. We extract the attention map of the \texttt{[CLS]} token from the self-attention layer of the model and obtain the variance of the attention map, which is exported along with the classification logits to assist in sorting the images, as shown in Figure \ref{ovr-frame}. Note that for ViT variants without \texttt{[CLS]} token (Swin), we replace the variance with the cumulative token of the attention map. 
After removing the unforgettable examples, the compressed training dataset $\hat{D}$ is described as:
\begin{equation}
\hat{D} = \{x_i| x_i \in D ,  f(x_i) \geq threshold\},
\end{equation}
where $f()$ stands for the “sort examples by forgetting counts” process. Therefore, $f(xi)$ is the list of examples $x_i$ sorted by counts. “$f(xi) \geq $ threshold” means choosing examples whose forgetting count is larger than the threshold. These selected examples will be our training subset $\hat{D}$.

\subsubsection{Online Example Filtering for Efficient Training}
\label{data_train}
Existing example sparsity works \cite{toneva2018empirical, yuan2021mest} on CNN split the training process into two phases: 1) Training with the complete dataset to obtain statistics on forgetting events. 2) Training with the remaining dataset by removing the less informative examples based on the learning statistics.
In this way, training acceleration can only be achieved in the second phase, which only accounts for 40\%$\sim$70\% of the total training process \cite{yuan2021mest}. This greatly limits the overall acceleration performance.

Unlike their semi-sparse training approach, we use an end-to-end sparse training approach, which keeps the training dataset sparse throughout the training process. 
This is the first time that the example sparsity is introduced in the ViT training scenario. Our approach is shown in the left part of Figure \ref{ovr-frame}.
The main idea is to first remove a random portion of training examples from the dataset, and then continuously update the remaining training subset during the training process by using a remove-and-restore mechanism. After several iterations, only relatively more informative examples remain in the training subset.


Specifically, before the training starts, we first reduce the training subset by randomly removing a given $m\%$ of training examples.
We define $r_e = 1 - m\%$ as the keep ratio of training examples.
Then, during the training, we periodically remove the $n\%$ least informative examples from the training subset ($n < m$) and randomly select $n\%$ examples from the removed examples to restore to the training subset.
And during the training process, we continuously track the number of forgetting events for each training example and identify the least informative examples.
In this way, the data set is optimized after several iterations during training, and $m\%$ the data size remains constant throughout the entire training process, leading to a more consistent training acceleration.

\subsection{Sparsity in Patches: Token and Attention Level}
\label{patch_sec}
The dense attention of ViTs leads to quadratic complexity with respect to the number of patches. 
Thus, introducing sparsity into patches can significantly reduce the computation. We consider both token-level sparsity and attention-level sparsity for patches based on the attention matrix.

\subsubsection{Revisiting ViT}
The ViT decomposes image $\mathbf{I}$ into $N$ non-overlapping tokens $ \{\mathbf{P}_i \}_{i=1}^N$, and apply a linear projection to obtain the patch embedding $\mathbf{X}_P \in \mathbb{R}^{N \times d}$, where $d$ is the dimension of embedding. 
The classification token $\mathbf{X}_{\text{cls}}$ is appended to accumulate information from other tokens and predict the class. Thus, the input to the transformer is $ \mathbf{X} = [\mathbf{X}_{\text{cls}}, \mathbf{X}_P] \in \mathbb{R}^{L \times d}$, where $L$=$N$+$1$ for short. 
The transformer encoder consists of a self-attention layer and a feed-forward network. In self-attention, the input tokens $\mathbf{X}$ go through three linear layers to produce the query ($\mathbf{Q}$), key ($\mathbf{K}$), and value ($\mathbf{V}$) matrices respectively, where $\mathbf{Q}$, $\mathbf{K}$, $\mathbf{V}$ $\in \mathbb{R}^{L \times d} $. The attention operation is conducted as follows:  
\begin{equation}
\text{Attention}(\mathbf{Q},\mathbf{K},\mathbf{V}) = \text{Softmax}( \mathbf{Q} \mathbf{K}^T/\sqrt{d}) \mathbf{V}.
\label{eq:self_attn}
\end{equation}
For multi-head self-attention (MHSA), $H$ self-attention modules are applied to $\mathbf{X}$ separately, and each of them produces an output sequence.

\subsubsection{Token-level Sparsity}
Previous works \cite{rao2021dynamicvit, kong2021spvit} suggests that pruning less informative tokens, such as the background, has little impact on the accuracy. Thus, we can improve training efficiency at the token level by eliminating redundant tokens. 
However, these works require additional token selector module to evaluate the importance of each token. Moreover, they target fine-tuning, which is incompatible with training from scratch, as additional losses need to be introduced. 
In this work, we aim to prune tokens without introducing additional modules and train ViTs from scratch with original training recipes.

In Eq. \ref{eq:self_attn}, $\mathbf{Q}$ and $\mathbf{K}$ can be regarded as a concatenation of $L$ token vectors: $\mathbf{Q} = [q_1, q_2,\ldots,q_L]^T $ and $\mathbf{K} = [k_1, k_2,\ldots,k_L]^T $. 
For the $i$-th token, the attention probability $ \mathbf{a}_i = \text{Softmax} ( q_i \cdot \mathbf{K}^T /\sqrt{d}) \in \mathbb{R}^{L}$ shows the degree of correlation of each key $k_i$ with the query $q_i$. 
The output $\mathbf{a}_i \cdot \mathbf{V} $ can be considered as a linear combination of all value vectors, with $\mathbf{a}_i$ being the combination coefficients. 
Thus, we can assume that $\mathbf{a}_i$ indicates the importance score of all tokens. Typically, a large attention value suggests that the corresponded token is important to the query token $i$.

For ViT, the final output only depends on the \texttt{[CLS]} token. Thus, the attention map of this special tokens $\mathbf{a}_{\text{cls}} = \text{Softmax}( q_{\text{cls}} \cdot \mathbf{K}^T / \sqrt{d}) $ represents the extent to which the token contributes to the final result. 
To this end, we utilize $\Tilde{\mathbf{a}}_{\text{cls}} \in \mathbb{R}^{N} $, which excludes the first element of $\mathbf{a}_{\text{cls}}$, as the criterion to select tokens. In MHSA, we use the average attention probability of all heads $\bar{\mathbf{a}_{cls}} = \frac{1}{H} \sum_{i=1}^{H} \Tilde{\mathbf{a}}^{(h)}_{\text{cls} }$. 
We select $K (K < N)$ most important patch tokens based on the value of $\bar{\mathbf{a}_{cls}}$ and define $R_T = \frac{K}{N} $ as the token keep ratio. Thus, only $1+R_TN$ tokens are left in the following layers.
%
%
%

For ViT variants without the \texttt{[CLS]} token, we calculate the importance scores of tokens by summing each column of the attention map \cite{wang2021spatten}. In detail, we use the cumulative attention probability $\sum_i \mathbf{a}_i \in \mathbb{R}^N$ to select informative tokens. We hypothesize that tokens that are important to many query tokens should be informative.


\subsubsection{Attention-level Sparsity }
Since attention maps are usually sparse, dense attention is unnecessary. Thus, we also introduce sparsity at attention level by pruning attention connections. In detail, given a query token, we only calculate its attention connections with a few selected tokens. Previous ViT variants usually utilize some hand-craft predetermined sparse patterns, such as a neighborhood window \cite{liu2021swin} or dilated sliding window \cite{beltagy2020longformer}. 
However, objects of different sizes and shapes may require different attention connections. Thus, these data-agnostic method is limited. Some input-dependent methods apply the deformable attention \cite{zhu2020deformable} to find corresponding tokens. However, these works also require additional modules to learn the selected tokens.

In this work, we utilize the attention vector of each patch token to decide critical connections. Thus, no additional modules are introduced.  
In detail, given the image patch token $\mathbf{P}_i$ and its corresponding query $q_i$, we use $\bar{\mathbf{a}_{i}}$ as the saliency criteria. 
In detail, we take the index of $R_A N$ tokens with the largest value of $\bar{\mathbf{a}}_i$, where $R_A \ll 1$ is the attention keep ratio. Thus, each patch token $\mathbf{P}_i$ only need to calculate the attention score with the selected $R_A N$ tokens for the rest layers.


\subsubsection{Token\&Attention Selector (TA-Selector) }
We further combine the above token selector and attention connections selector into a single module, named \textit{TA-Selector}. 
As shown in the right part of \cref{ovr-frame}, the TA-Selector is inserted after the self-attention module of one transformer encoder layer. 
Given the attention matrix $\mathbf{A} = \text{Softmax} ( \mathbf{Q} \mathbf{K}^T/\sqrt{d})\in \mathbb{R}^{L\times L}$, we first utilize the attention probability of $\texttt{[CLS]}$ tokens $\mathbf{a}_{\textbf{cls}}$ (the blue row of \cref{ovr-frame}) to locate $R_TN$ most informative tokens. We denote the index of selected tokens as set $\mathcal{T}$. 
We then extract the attention map of select tokens $ \mathbf{A}_{\mathcal{T}}$, whose element is defined by $\mathbf{A}[i,j]$ with $\{i,j\}\in \mathcal{T} \times \mathcal{T}$ (the ticked red cell). Then we perform the attention-selector on $ \mathbf{A}_{\mathcal{T}}$. 
In summary, $1+R_TN$ tokens are left after the TA-selector, and each token only calculates the attention  with $R_T R_A N$ tokens in the following transformer encoder layers. 

With TA-Selector, we dynamically prune both tokens and attention connections at the same time.  
As the attention map at early layer may be inaccurate \cite{xu2021evo}, we follow a hierarchical pruning scheme, which incorporate the TA-Selector between transformer encoder layers several times. Thus, we gradually prune tokens and attention connections with the network going deeper.

\subsection{Sparsity Determination of Each Level}
We focus on exploring the maximum compression sparsity (ratio) for each level while preserving accuracy. We leverage a hierarchical data redundancy reduction scheme by setting the sparsity one by one in order: example $\to$ token $\to$ attention.
This is based on two reasons: the characteristics of each level and the training efficiency.
The example level is relatively independent, there is no sparsity trade-off between the example level and the TA level.  Example level sparsity ratio contributes directly and the most to the time saving, so it should be considered first. After that, token and attention.
When performing token-attention, we also prioritize pruning tokens over attention. There are two reasons: 1). Token contributes more speed-up compared to attention even with a small keep ratio (Set the attention keep ratio $R_A$ to 0.2 with 4.8\% speed up in Table \ref{tab:sub_method}). 2). Pruning attention is less sensitive to accuracy, which means it can be used to squeeze out the redundancy from a finer perspective after performing example and token sparsity.
There are more complex ways to determine the sparsity, such as by adding constraints during training, which may add additional computation costs to training.
We first run each level with a different keep ratio to see the performance degradation trend (0.9,0.8,0.7, etc.). Then, these results can be considered as a look-up table to guide the decision of Tri-level sparsity.

\section{Experiments}
\subsection{Datasets and Implementation Details}
Our experiments are conducted on ImageNet-1K \cite{deng2009imagenet}, with approximately 1.2 million images. We report the accuracy on the validation set with 50k images. The image resolution is $224 \times 224$. We also report results on Imagenet-V2 matched frequency \cite{recht2019imagenet} to control overfitting. 
We apply our framework to different backbones including  DeiT~\cite{Touvron2021TrainingDI}, Swin~\cite{liu2021swin}, PiT~\cite{heo2021pit}, and LV-ViT~\cite{NEURIPS2021_9a49a25d} with the corresponding settings.
 In detail, the network is trained for $300$ epochs on 4 NVIDIA V100 GPUs and is optimized by AdamW~\cite{loshchilov2018decoupled} with weight decay $0.05$. The batch size is set to $512$, The learning rate is set to $5 \times 10^{-4}$ initially, and is decayed with a cosine annealing schedule. 
For example level sparsity, we apply the remove-and-restore approach every $30$ epochs with 5\% data removal for each iteration.
Since this approach takes extra time and the accuracy decreases slightly after each iteration, we only iterate the number of times in which all the removed data can be covered.
Hence, we set the different number of iterations for different keep rates.
For 10\% removed data, we iterate 3 times (30th, 60th, 90th epoch). For 20\% removed data, we iterate 5 times (30th, 60th, 90th, 120th, 150th epoch).
For token and attention level sparsity, we insert our TA-Selector before the $4^{th}$, $7^{th}$, $10^{th}$ layer for hierarchical pruning scheme as in \cite{rao2021dynamicvit}. Besides, during the training process, we adopt a warm-up strategy for the TA-Selector. The token keep ratio $R_T$ starts from $1.0$ and is gradually reduced to the given $R_T$ with a cosine annealing schedule.  
For simplicity and fair comparison, we set keep ratio to $0.7$, $0.8$, and $0.9$ in our main experiments, denoting as Tri-Level/0.7, Tri-Level/0.8 and Tri-Level/0.9, respectively.

\begin{table}[t]
\begin{center}
\begin{small}
\begin{sc}
\resizebox{1.0\columnwidth}{!}{
\begin{tabular}{l|cccc}
\toprule
\multirow{2}{*}{Method}  & Training   & MACs (G)      & 1K   &  V2  \\
      & Time Reduced    &  Saving    & Acc(\%)   & Acc(\%) \\
\midrule
 \multicolumn{5}{c}{Deit-T}  \\
\midrule
Dense  & - & -  & 72.2 & 65.0 \\
DynamicViT/0.7 \cite{rao2021dynamicvit} &  0\% &  30.0\% & 71.8 (-0.4) & 64.6 \\
S$^{2}$VITE \cite{chen2021chasing}  & 10.6\% &  23.7\% & 70.1 (-2.1)  & -\\ 
\rowcolor{light-gray} 
Tri-Level/0.9   & 15.2\%  &  17.4\%  &  \textbf{72.6 (+0.4)} & \textbf{65.9} \\
\rowcolor{light-gray} 
Tri-Level/0.8   & 29.4\%  &  31.9\%  & 72.1 (-0.1) &65.0\\
\rowcolor{light-gray} 
Tri-Level/0.7  &  35.6\%  &  38.5\% & 71.9 (-0.3) & 64.6\\
\midrule
\multicolumn{5}{c}{DeiT-S}  \\
\midrule
Dense  & - & -  & 79.8  & 73.6 \\
IA-RED$^2$ \cite{pan2021iared2}  & 0\% & 32.0\%  & 79.1 (-0.7) & - \\
DynamicViT/0.7 \cite{rao2021dynamicvit} &  0\% &   36.7\% & 79.3 (-0.5) &72.8 \\
UVC \cite{yu2022unified} & 0\% & 42.4\% & 79.4 (-0.4) & - \\
DynamicViT/0.7* \cite{rao2021dynamicvit}  & 7.2\% &  34.7\% & 77.6 (-2.2) &71.1 \\
EViT/0.8 \cite{liang2022evit}  &  6.0\% &  13.0\% & 79.8 (-0.1) & 73.4  \\
EViT/0.7 \cite{liang2022evit}  &  20.0\% &  35.0\% & 79.5 (-0.3) & 73.2  \\
EViT/0.6 \cite{liang2022evit}  &  25.0\% &  43.0\% & 78.9 (-0.9) & 72.3  \\
S$^{2}$VITE \cite{chen2021chasing}  & 22.7\% &  31.6\% & 79.2 (-0.6) & - \\
\rowcolor{light-gray} 
Tri-Level/0.9   &  15.7\%  &  17.0\% & \textbf{79.9 (+0.1)} & \textbf{73.8}\\
\rowcolor{light-gray} 
Tri-Level/0.8  &  29.8\%  &  31.3\% & 79.5 (-0.3)  & 73.3\\
\rowcolor{light-gray} 
Tri-Level/0.7   & 34.3\%  &  37.6\%  & 79.3 (-0.5)  & 72.9\\
\midrule
 \multicolumn{5}{c}{DeiT-B}  \\
\midrule
 Dense & - & -  & 81.8 & 75.8\\
IA-RED$^2$  \cite{pan2021iared2}  & 0\% & 33.0\%  & 80.3 (-1.5) & -\\
DynamicViT/0.7 \cite{rao2021dynamicvit}  &  0\% &  37.4\% & 80.7 (-1.1) & 74.2\\
EViT/0.7 \cite{liang2022evit}  &  15.2\% &  35.0\% &  81.3 (-0.5) & 75.1\\
\rowcolor{light-gray} 
Tri-Level/0.9   &  10.2\% & 15.6\%  & \textbf{81.6 (-0.2)} & \textbf{75.3}\\
\rowcolor{light-gray} 
Tri-Level/0.8  &  23.7\% &  29.1\% & 81.3 (-0.5) & 75.1 \\
\midrule
 \multicolumn{5}{c}{Swin-T}  \\
\midrule
 Dense & - & -  & 81.2 & 75.1 \\
 DynamicSwin/0.9	 \cite{rao2022dynamic}  &  0\% &  7\% & 81.0 (-0.2) & 75.0 \\
\rowcolor{light-gray} 
Tri-Level/0.9   &  13.1\% &  15.6\% & \textbf{81.2 (-0.0)} & \textbf{75.1}  \\
\rowcolor{light-gray} 
Tri-Level/0.8  &  21.1\% &  25.3\% & 81.0 (-0.1) & 74.9\\
\midrule
 \multicolumn{5}{c}{Swin-S}  \\
\midrule
Dense & - & -  & 83.2 & 76.9  \\
DynamicSwin/0.7	 \cite{rao2022dynamic}  &  0\% & 21\% & 83.2(-0.0) & 76.9 \\
WDPruning \cite{yu2022width} & 0\% & 12.6\%  &  82.4 (-0.6) & 76.1 \\
\rowcolor{light-gray} 
Tri-Level/0.8   &  20.2\% & 24.1\%  & \textbf{83.2 (-0.0)} & 76.8 \\
\bottomrule
\end{tabular}
}
 \caption{The training time reduced, MACs (G) saving, and Top-1 accuracy comparison on the ImageNet-1K and Imagenet-V2 matched frequency.  “*” refers to our reproduced results of DynamicViT training from scratch.}
 \label{main_table}
\end{sc}
\end{small}
\end{center}
\end{table}

\subsection{Experimental Results}
We report the Top-1 accuracy of each model with different keep ratio. For efficient training metrics, we evaluate the reduced running time and MACs (G) saving of the entire training process.
Our main results are shown in \cref{main_table}. Remarkably, our efficient training method can even improve the ViT accuracy rather than compromising it (15.2\% time reduction with 72.6\% Top-1 accuracy on Deit-T, and 15.7\% time reduction with 79.9\% Top-1 accuracy on Deit-S). This demonstrates the existence of data redundancy in ViT.

We also compare our method with other efficient ViTs aiming at reducing redundancy. Most of the existing efforts target only efficient inference. They either apply to the fine-tuning phase \cite{rao2021dynamicvit, pan2021iared2} or introduce additional modules with additional training cost \cite{chen2021chasing, yu2022unified}.
In contrast to others, we can accelerate both the training and inference process. Our method significantly reduces the training time while restricting the accuracy drop in a relatively small range. Notably, we are able to reduce the training time by $35.6\%$ for DieT-T with a negligible $0.3\%$ accuracy degradation, and $4.3\%$ for DieT-S with a $0.5\%$ decrease in accuracy, which outperforms existing pruning methods in terms of accuracy and efficiency. 


\subsection{Generalization on other Architectures and Tasks}
To verify the generalizability of our approach, we conduct experiments with other vision transformer models and evaluate them on object detection and instance segmentation tasks.

\subsubsection{Swin Transformer}
We further apply our method to other ViT variants, such as Swin Transformer \cite{liu2021swin}. Instead of using the \texttt{[CLS]} token, it applies a global average pooling on features of all tokens. Thus, we use the cumulative attention probability to select tokens. 
For the example level sparsity, we also replace the variance of \texttt{[CLS]} token's attention map with that of the cumulative attention map. 
%
%
%
%
%
Results are shown in Table \ref{main_table}. We are able to reduce the training time by $13.1\%$ for Swin-T and $20.2\%$ for Swin-S with no accuracy drop. Thus, our method can generalize to other ViT architectures. 
More results for other architectures such as LV-ViT \cite{NEURIPS2021_9a49a25d} and other datasets such as ImageNet-Real and CIFAR can be found in the Appendix.
\subsubsection{Object detection and Instance Segmentation}
Experiments are conducted on COCO 2017 \cite{lin2014microsoft}, which contains 118K training, 5K validation and 20K test-dev images. As shown in Table \ref{tab:obj_dec}, we use the Mask R-CNN framework and replace different backbones. FLOPs are measured at 800 × 1280 resolution. Our Tri-level model denotes the 13.1\% training time reduction model in Table \ref{main_table}. 

\begin{table}[t]
		\centering
            \resizebox{1.0\columnwidth}{!}{
            \begin{tabular}{l|ccc|ccc}
            \toprule
            Backbone & $AP^b$ & $AP^b_{50}$  & $AP^b_{75}$ & $AP^m$ & $AP^m_{50}$ & $AP^m_{75}$ \\
            \midrule
            ResNet50 \cite{he2016deep} & 41.0 &61.7  &44.9 & 37.1  &58.4  &40.1 \\
            PVT-S \cite{wang2021pyramid} & 43.0&65.3&46.9&	39.9&62.5&42.8  \\
            Swin-T & 46.0	&68.1	&50.3&	41.6	&65.1	&44.9 \\
            \rowcolor{light-gray} 
            Tri-level  & 46.0	&68.0	&50.3&	41.6&	65.2	&44.9  \\ 
            \bottomrule
            \end{tabular}}
            \caption{Results on COCO object detection and instance segmentation.}
            \label{tab:obj_dec}
\end{table}

\subsection{Deployment on Edge Devices}
We also evaluate our method on an embedded FPGA platform, namely, Xilinx ZCU102. The platform features a Zynq UltraScale + MPSoC device (ZU9EG), which contains embedded ARM CPUs and 274k LUTs, 2520 DSPs, and 32.1Mb BRAMs on the programmable logic fabric. The working frequency is set to 150MHz for all the designs implemented through Xilinx Vitis and Vitis HLS 2021.1.
The 16-bit fixed-point precision is adopted to represent all the model parameters and activation data. 
We measure both on device training latency and inference latency. 
In detail, the training latency is measured on FPGA using a prototype implementation that supports layer-wise forward and backward computations. We report the average per epoch training time. Both inference and training use the batch size of 1. 
As in Table \ref{tab:latency}, our method can also accelerate the training and inference on edge devices. 


\begin{table}[t]
        
		\centering
		\resizebox{0.78\columnwidth}{!}{
            \begin{tabular}{l|ccc}
            \toprule
            \multirow{2}{*}{Method}  & MACs (G)  & Training   & Inference \\ 
                  & Saving & Latency (hour)    & Latency (ms)  \\
            \midrule
             \multicolumn{4}{c}{DeiT-T}  \\
            \midrule
            Dense      & - &  13.79 & 10.91 \\
           \rowcolor{light-gray} Tri-Level/0.9  & 17.4\% & 11.49 &  9.68\\
           \rowcolor{light-gray} Tri-Level/0.8  & 31.9\% & 9.61 & 8.50 \\
           \rowcolor{light-gray} Tri-Level/0.7  & 38.5\% & \textbf{8.91} & \textbf{7.36}\\
            \midrule
             \multicolumn{4}{c}{Deit-S}  \\
            \midrule
            Dense       & -   & 32.44 & 25.68 \\
          \rowcolor{light-gray}  Tri-Level/0.9  & 17.0\% & 27.28 & 22.92 \\
          \rowcolor{light-gray}  Tri-Level/0.8  & 31.3\%  & 22.55 & 20.25\\
          \rowcolor{light-gray}  Tri-Level/0.7  & 37.6\% & \textbf{21.15} &  \textbf{17.61}  \\
            \bottomrule
            \end{tabular}}
\caption{The inference latency and the average per epoch training latency on Xilinx ZCU102 FPGA board.}
\label{tab:latency}
 \end{table}
 
\begin{table}[t]
\centering
\resizebox{0.9\columnwidth}{!}{
\begin{tabular}{l|lcccc}
\toprule
\multirow{2}{*}{Level}&  \multirow{2}{*}{Method} & Keep & Training     & 1K & V2  \\
&  & Ratio  & Time Reduced   & Acc.(\%)  & Acc.(\%) \\
\midrule
& Dense      & 1.0 & -   & 79.8 & 73.6 \\
\midrule
\multirow{6}{*}{Example}   &  Random & 0.8 & 19.9\% & 78.4 & 71.7\\
&  Active Bais & 0.8 & 4.3\% & 79.1 & 72.5 \\
&   Forgetting &  0.8 & 16.1\% & 79.0 & 72.4 \\
&  \textbf{ours}  & 0.9 &  9.9\%    & \textbf{80.1}   & 73.8 \\
&  \textbf{ours}    & 0.8 &  19.8\%    & 79.7            & 73.5 \\
&   \textbf{ours}  & 0.7 &  29.8\%    & 79.2  & 72.6 \\
\midrule
\multirow{7}{*}{Token}    &  Static & 0.7 & 20\% & 75.2 & 68.3\\
&  DynamicViT & 0.7 & 0\% & 79.3 & 72.8\\
&  IA-RED$^2$ & 0.7 & 0\% & 79.1 & -\\
&  Evo-ViT & 0.7 & 13.0\% & 79.4 & -\\ 
&  \textbf{ours}  & 0.9 &   6.2\%    & \textbf{79.9} & 73.8 \\
&  \textbf{ours}  & 0.8 &    13.3\%  & 79.7 & 73.4 \\
&   \textbf{ours} & 0.7 &    20.6\%  & 79.4 & 73.2 \\
\midrule
\multirow{5}{*}{Attention}  &   Magnitude & 0.5 & 2.8\% & 78.9 & 72.3\\
&  Longformer & 0.2 & 4.8\% & 78.8 & 72.0 \\
& \textbf{ours}     & 0.4 &  3.3\% & \textbf{79.8}  & 73.5  \\
& \textbf{ours}  &  0.3 &    3.6\%     & 79.7  & 73.5 \\
& \textbf{ours}   & 0.2  &    4.8\%    & 79.7  & 73.4  \\
\bottomrule 
\end{tabular}
}
\caption{Sub-method effectiveness on DeiT-S under different keep ratios. We also compare with existing individual example/token/attention sparsity methods.}
\label{tab:sub_method}
\end{table}

\vspace{-0.5em}
\section{Ablation Study}
\subsection{Sub-method Effectiveness}
We conduct ablation studies to evaluate the contribution of each sparse method separately. 
The results of DeiT-S with different keep ratios at each sparse level are shown in \cref{tab:sub_method}. Removing redundant data in each sub-method can even lead to improvements in accuracy while reducing training time.
At the same keep ratio, the example level sparsity contributes the most to the training time reduction, as it directly removes the entire image. Thus, it can be considered the coarse-grained level of data reduction. Meanwhile, the attention level has less reduction in training time even at low keep ratios because it only contributes to the matrix multiplication of the MHSA module. Thus, it can be considered a fine-grained level of data reduction.

\subsection{Comparison of Different Methods}
To verify the advantages of our method, we compare it with other data sparsity strategies on DeiT-S. As shown in Table \ref{tab:sub_method}, \textit{Random} denotes randomly updating the training data with the same frequency and number of updates as our method. Under the same keep ratio, it shows a significant accuracy degradation (78.4\% vs. 79.7\% ). We also compare it with the existing work \cite{toneva2018empirical}, which conducts a two-phase method for example evaluation and removal. The accuracy is lower (79.0\% vs. 79.7\%) and the training acceleration is limited ($16.1\%$ vs. $19.8\%$). 



\subsection{Visualization}
\begin{figure}[t]
\vskip -0.2in
\begin{center}
\centerline{\includegraphics[width=0.8 \columnwidth]{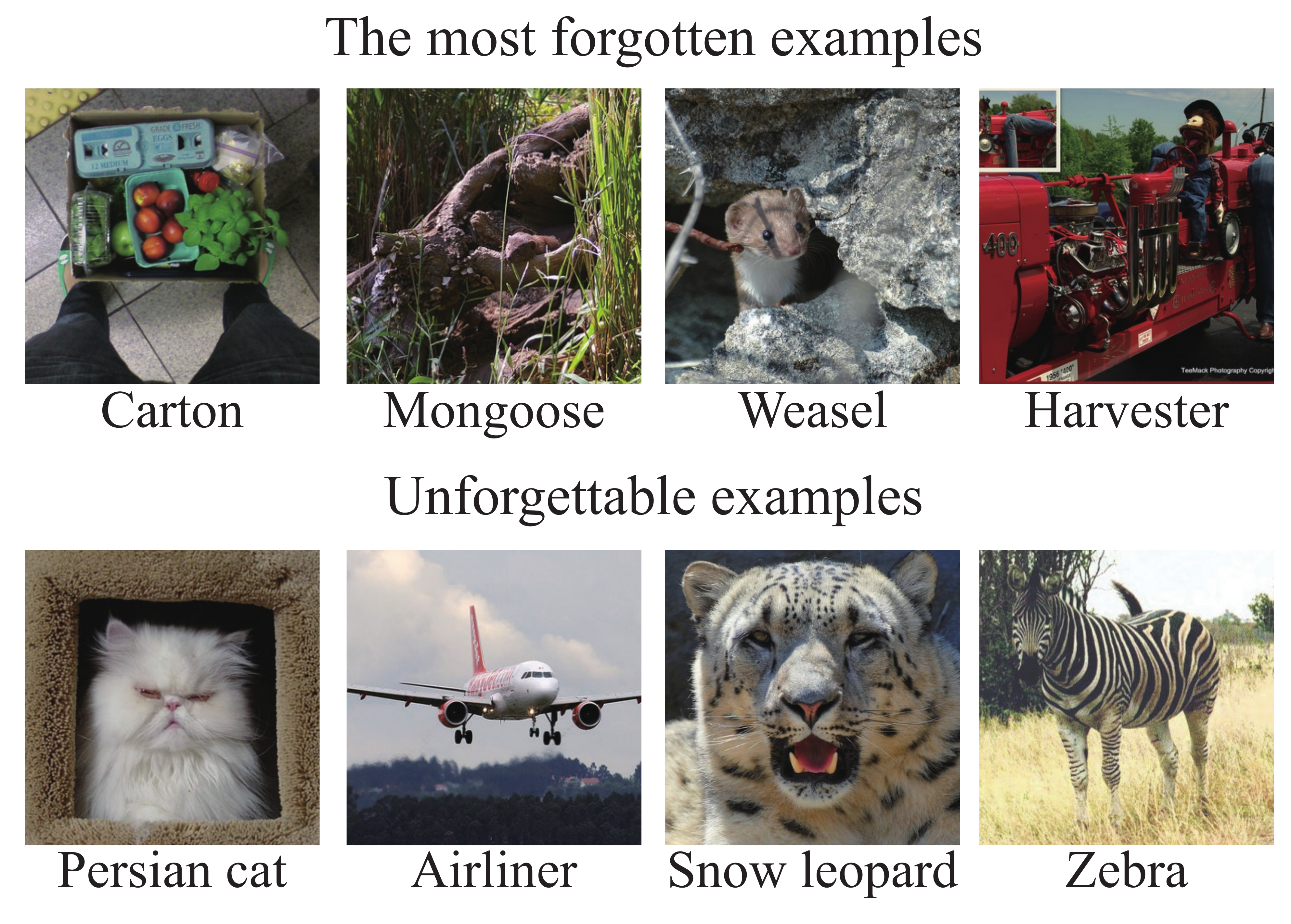}}
\caption{Visualization of the most forgotten examples and unforgettable examples.}
\label{fig:visual_examples}
\end{center}
\end{figure}

\noindent\textbf{Example Evaluation.} 
We visualize both of the most forgotten (hardest) training examples and the unforgettable (easiest) training examples obtained by our proposed online example filtering method.
As shown in \cref{fig:visual_examples}, the hardest examples generally contain more sophisticated content in the image, making them more informative for training.
On the contrary, the less informative examples are usually easier to be recognized, and they contain conspicuous characteristics of their categories.
This observation is similar to the previous works~\cite{toneva2018empirical,yuan2021mest}, while our online example filtering method can take the advantage of data efficient training throughout the entire training process.


\begin{figure}[t]
\vskip -0.2in
\begin{center}
\centerline{\includegraphics[width=0.8 \columnwidth]{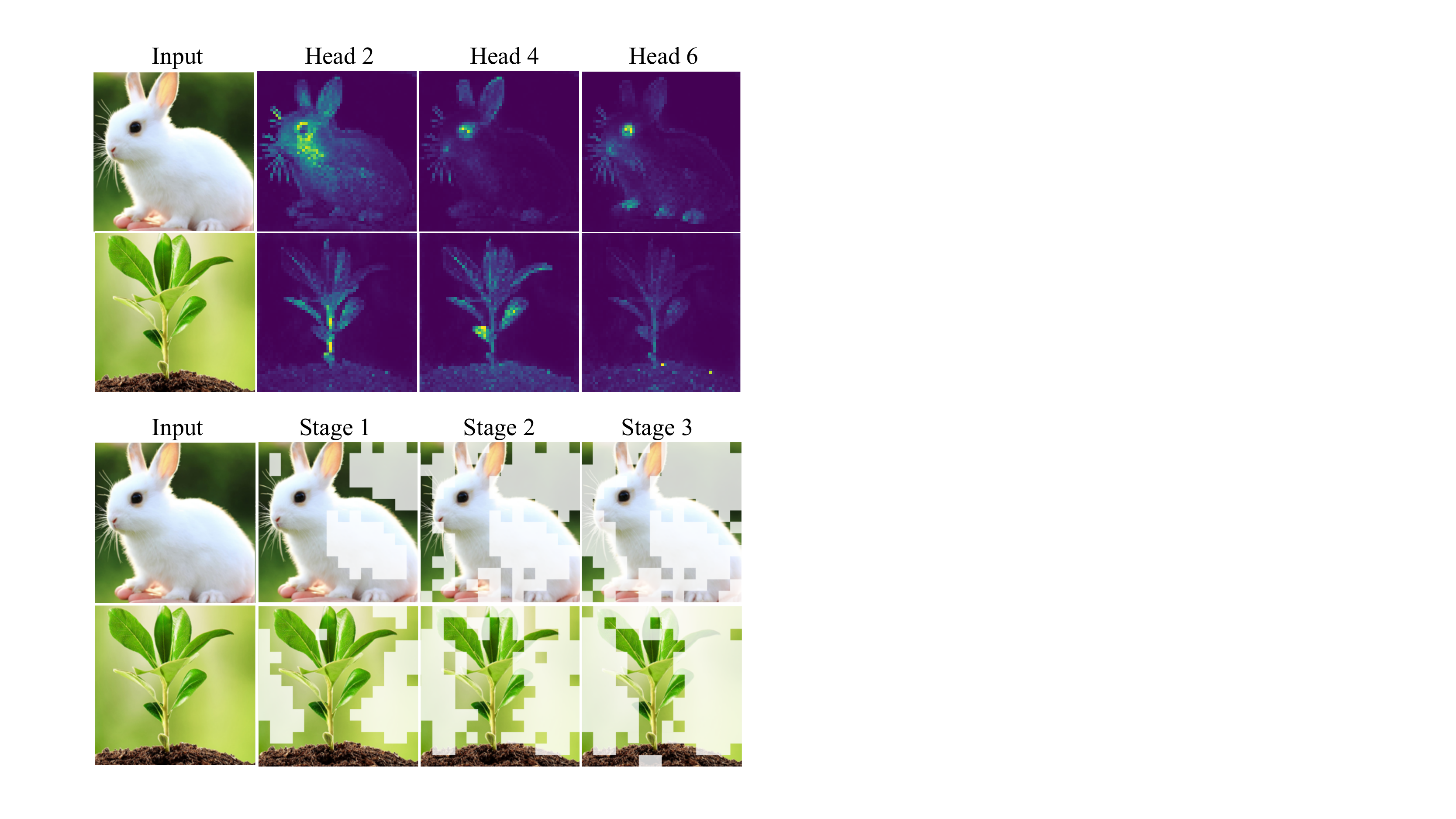}}
\caption{Visualization of token pruning. The masked patches represent the less informative tokens that are pruned.}
\label{fig:select_token}
\end{center}
\vskip -0.2in
\end{figure}

\noindent\textbf{Token Selection.}
We also visualize the tokens identified by our attention-based token selector in Figure \ref{fig:select_token}. Note that the network can gradually remove the background tokens, which is less informative. Eliminating these redundant tokens can reduce the computation at both training and inference time. Moreover, the remaining tokens maintain class-specific attributes, which is helpful for recognition.

\section{Conclusion}
\label{conclusion}
In this work, we introduce sparsity into data and propose an end-to-end efficient training framework to accelerate ViT training and inference. We leverage a hierarchical data redundancy reduction scheme, by exploring the sparsity of the number of training examples in the dataset, the number of patches (tokens) in each input image, and the number of connections between tokens in attention weights. Comprehensive experiments validate the effectiveness of our method. Future work includes extending our framework to other aspects such as weight redundancy. 

\newpage
\bibliography{template} 

\newpage
\clearpage

\appendix

{\LARGE \bf Appendix}
\section{Definition of Example Forgetting}
\label{def_forget}
Some definitions for the example forgetting approach applied in our example selector are as follows:
\begin{itemize}[leftmargin=*, noitemsep,topsep=0pt]
    \item \textbf{Forgetting Events}: A training example learned in the previous training process is being mis-classified in latter training process, in other words, forgotten by the network.
    \item \textbf{Learning Events}: A training example mis-classified in the previous training process is being classified correctly in the latter training process.
    \item \textbf{Unforgettable Examples}: A training example after learned by the network  at a certain step will not be mis-classified (forgotten) during the rest of the training course.  
\end{itemize}
According to prior work \cite{toneva2018empirical}, the unforgettable examples are generally considered easy to be learned and less informative than others, so they can be removed based on the order of the number of forgotten events. Networks trained on the sub-dataset can still maintain generalization performance compared to networks trained on the full dataset.

We choose the forgetting score as an evaluation metric over other methods for the following reasons:
\begin{itemize}[leftmargin=*, noitemsep,topsep=0pt]
\item \textbf{Better fit for our online examples filtering method}:
The forgetting approach counts the number of times an example is correctly classified or not (forgotten). Examples that have 0 or low forgetting counts are defined as simple examples, and thus can be removed. Therefore, it is a direct way to guide removing examples.
\item \textbf{Better for efficient training}:
Our method does not involve reweighting all the examples \cite{chang2017active}  or having interaction with the loss \cite{katharopoulos2018not}. Our evaluation method is relatively independent of training, by counting the output logits misclassified times. So it can easily be applied to various models with different training schemes to improve training efficiency.
\item \textbf{More suitable for combining with ViT}:
The forgetting approach uses the output logit of the model. It aligns with calculating the variance of the attention map of the [CLS] token or the sum of each column of the attention map in our proposed method.
\end{itemize}

\section{Explore Example Level  Sparsity}
\subsection{Different Update Frequency.}
We experimented with the update frequency and number of updates for the data, and the results are shown in Table~\ref{sample-table}. We keep the same training epochs as DeiT, and train with 80\% of the dataset. We apply the  remove-and-restore approach with 5\% data during each update. We first set the update frequency to once every 30 epochs, with a total of 10 times. However, the accuracy is only 78.7\%. This is because when the data is updated, the training accuracy will descend due to the introduction of new data, and it takes 5-10 epochs to recover the accuracy. Therefore, the accuracy is low under the restriction of 300 epochs. We reduce the number of updates to only 5 updates, and no updates after 150 epochs to reduce the loss from updates. Since 20\% of the data is removed and 5\% is updated each time, 5 times is enough to iterate 20\% of the data.  In addition, the training will converge faster because the quality of data becomes better. The latter 150 epochs can be regarded as a phase of retrain. Under this setting, the accuracy reaches 79.7\%.

\begin{table}[t]
		\centering
            \resizebox{0.8\columnwidth}{!}{%
            \begin{tabular}{l|ccc}
            \toprule
            \multirow{2}{*}{Method}  & Keep    & Top-1   & Top-5\\
                  & Ratio   & Acc.(\%) & Acc.(\%) \\
            \midrule
            No update           & 0.8   & 78.9  &94.4\\
            \midrule
            30 epoch 10 times   & 0.8     & 78.7 &94.3 \\
            30 epoch 5 times    & 0.8     & \textbf{79.7}   &\textbf{94.8}   \\
            40 epoch 5 times    & 0.8     & 78.6 &94.3 \\
            40 epoch 4 times    &  0.8     & 79.0  &94.5 \\
            \bottomrule
            \end{tabular}}
             \caption{Different update frequency and number of updates for the example level on DeiT-S.}
            \label{sample-table}
\end{table}

\subsection{Shuffle vs. Non-Shuffle}
As mentioned in the main paper, we remove a portion of examples from the dataset prior to training. Thus, when the dataset is updated during training, the newly removed data will be put into the existing pool of removed examples, and then the restored data will be retrieved from here as well. We found that shuffling the data in the pool before the restore operation can effectively improve the training accuracy. As shown in \cref{data_bal}, shuffling the data before restoring can result in a 0.3\% accuracy increase compared to not shuffling (79.7\% vs. 79.4\%). Under the $0.9$ keep ratio, our proposed example level sparsity method can even improve the Deit-S accuracy by $0.3\%$ rather than compromising it.

\subsection{Class Balance vs. Class Imbalance}
When we perform remove-and-restore operations on the examples to update the training sub-dataset, there are two ways: (i) Class balance approach, which removes the same percentage of data for each class according to the forgetting events. 
(ii) Class imbalance approach, which calculates the forgetting events uniformly for all the examples, and then removes a certain percentage of the least forgettable data, regardless of the class label of the example. 
The results of both approaches are shown in \cref{data_bal}. Class imbalance outperforms class balance method by 1.0\%  (79.4\% vs. 78.4\%) under 0.8 keep ratio on DeiT-S. This is due to the fact that different classes in the dataset have different levels of difficulty in training. If all classes remove data using the same ratio, then for difficult classes, forgettable examples may be removed, causing a decrease in training accuracy.

\begin{table}[h]
\begin{center}
\begin{small}
\begin{sc}
\resizebox{0.85\columnwidth}{!}{
\begin{tabular}{lccc}
\toprule
\multirow{2}{*}{Method}  & Keep     & Top-1  & Top-5\\
      & ratio  & Acc.(\%)  & Acc.(\%)\\
\midrule
Full Data        & 1.0   & 79.8 & 94.9 \\
\midrule
Class Balance    & 0.8     & 78.4 &94.2 \\
Imbalance           & 0.8     & 79.4  &94.7 \\
Imbalance + Shuffle & 0.8      & 79.7 &94.8 \\
Imbalance + Shuffle & 0.9    & \textbf{80.1}  &\textbf{95.0} \\
\bottomrule
\end{tabular}
}
\caption{Comparison of different example level sparsity strategy on DeiT-S.}
\label{data_bal}
\end{sc}
\end{small}
\end{center}
\end{table}

\subsection{[CLS] Attention Map Selection}
For ViTs with [CLS] tokens, we choose the attention map of [CLS] from different layers to evaluate the complexity of the images, and conduct experiments for both example level and Tri-level.
As shown in Table \ref{cls_sel}, when running only the example level, using the attention map of the last layer can achieve the highest accuracy, as it is the most accurate for evaluating the importance of the token.
When running Tri-level, the [CLS] token at the last layer contains less information due to the combination of token and attention level sparsity. Therefore, we choose the attention map of the third layer, which is the layer before token pruning to feed into our example selector. The accuracy is higher than using the last layer (72.6\% vs. 72.4\%).

\begin{table}[h]
\begin{center}
\begin{small}
\begin{sc}
\resizebox{0.85\columnwidth}{!}{
\begin{tabular}{l|ccc}
\toprule
\multirow{2}{*}{Method}  & Layer Number    & Top-1  & Top-5\\
     &   & Acc.(\%)  & Acc.(\%)\\
\midrule
\multirow{3}{*}{Example Level}   & w/o [CLS]   & 72.4     & 91.4\\
                                  & 3     & 72.8     &91.5 \\
                                 & 12      & \textbf{73.0}    & \textbf{91.5}\\
\midrule
\multirow{3}{*}{Tri-Level}     & w/o [CLS]       &  72.2  &  91.1    \\
                           & 3        &  \textbf{72.6}   & \textbf{91.5}\\
                           & 12    &  72.4 &  91.4   \\
\bottomrule
\end{tabular}
}
\caption{Comparison of different attention map selection on DeiT-T at 0.9 keep ratio.}
\label{cls_sel}
\end{sc}
\end{small}
\end{center}
\end{table}

\section{Generalization on others Architectures and Datasets}
\subsection{Implementation Details}
We describe the token pruning locations for Swin \cite{liu2021swin}, \cite{NEURIPS2021_9a49a25d} and PiT \cite{heo2021pit}.
Swin has 4 stages, the configurations are: Swin-T: C = 96, layer numbers = \{2, 2, 6, 2\}; Swin-S: C = 96, layer numbers = \{2, 2, 18, 2\}
The Patch Merging layer between each stage makes the token non-transitive through stages. Also, the 1,2,and 4 stage has only one pair of W-MSA and SW-MSA transformers, making it hard to prune tokens. Therefore, we perform token pruning on stage 3, with the most number of layers. In side stage 3, we prune token after the \{1,2,3\} layer for Swin-T, and \{2,4,6\} layer for Swin-S, for fair comparison against \cite{rao2022dynamic}.
For LV-ViT-S, we insert the token selector after the \{4, 8, 12\} layers. For LVViT-M, we insert the token selector after the \{5, 10, 15\} layers. For PiT-XS/S, we insert the token selector after the \{1, 5, 10\} layers, for fair comparison against \cite{rao2021dynamicvit,kong2021spvit,liang2022evit}.

\subsection{Results on Other Datasets}
We train on four datasets on DeiT-S: Imagenet Real \cite{deng2009imagenet}, CIFAR-10 \cite{krizhevsky2009learning}, CIFAR-100 \cite{krizhevsky2009learning}, and Flower \cite{nilsback2008automated} under the same hyperparameters. We evaluate the accuracy under different keep ratios: 1.0, 0.7, and 0.5. Results are shown in Table~\ref{tab:other_dataset}. All the standard deviation of each sub-method is about  $-0.2 \sim +0.2$
\begin{table}[t]
\begin{center}
\begin{small}
\begin{sc}
\resizebox{1.0\columnwidth}{!}{
\begin{tabular}{ccccc}
\toprule
Keep  & Real & CIFAR-10    &  CIFAR-100   &  Flowers \\
Ratio    & $(\% )$ & $(\% )$    &  $(\% )$   &  $(\% )$  \\
\midrule
1.0     &  85.7          &  98.0         & 87.1        & 98.8 \\
0.7     &  85.3±0.1 (-0.4)   &  97.9±0.0 (-0.1)   & 87.0±0.1 (-0.1) & 98.8±0.0 (-0.0) \\
0.5     &  84.2±0.2 (-1.5)   &  97.2±0.0 (-0.8)  & 86.1±0.1 (-1.0)  & 98.7±0.0 (-0.1)\\
\bottomrule
\end{tabular}
}
\caption{Example-level sparsity of different datasets on DeiT-S.}
\label{tab:other_dataset}
\end{sc}
\end{small}
\end{center}
\vskip -0.2in
\end{table}

\subsection{Results on Other ViT Variants}
We apply our proposed method to LV-ViT \cite{NEURIPS2021_9a49a25d} and PiT \cite{heo2021pit}, which outperforms state-of-the-art compression methods in terms of training efficiency and accuracy, as shown in Table \ref{lvvit_res}. We also show the results on the COCO dataset in Table \ref{tab:pit_coco}. We use the same settings as the \cite{heo2021pit}, where we replace the backbones of the Deformable DETR \cite{zhu2020deformable} and reduce the image resolution for training. Our model can reduce 19.3\% training time without sacrificing any performance on both classification and object detection tasks compared to PiT-S.

\begin{table}[h]
\begin{center}
\begin{small}
\begin{sc}
\resizebox{1.0\columnwidth}{!}{
\begin{tabular}{l|ccc}
\toprule
\multirow{2}{*}{Method}  & Training   & MACs (G)      & Top-1  \\
      & Time Reduced    &  Saving    & Acc(\%) \\
 \midrule
 \multicolumn{4}{c}{PiT-XS}  \\
 \midrule
 Dense  & - & -  & 78.1 \\
 SPViT \cite{kong2021spvit}  & 0\% &  18.6\%  & 78.0 (-0.1) \\
\rowcolor{light-gray} Tri-Level   & 24.1\% &  28.6\%  & \textbf{78.1 (-0.0)} \\
  \midrule
 \multicolumn{4}{c}{PiT-S}  \\
 \midrule
 Dense  & - & -  & 80.9 \\
 SPViT \cite{kong2021spvit}  & 0\% & 11.0\% & 80.9 (-0.0) \\
\rowcolor{light-gray} Tri-Level   & 19.3\% & 20.1\%  & 80.9 (-0.0) \\
 \midrule
  \multicolumn{4}{c}{LV-ViT-S}  \\
\midrule
Dense  & - & -  & 83.3 \\
DynamicViT \cite{rao2021dynamicvit} &  0\% &  30.2\% & 83.0  (-0.3)  \\
EViT \cite{liang2022evit}  & 15.0\% & 29.0\% & 83.0 (-0.3)  \\
\rowcolor{light-gray} 
\rowcolor{light-gray} Tri-Level  & 23.6\%  & 26.7\% & \textbf{83.1 (-0.2)}  \\
 \midrule
 \multicolumn{4}{c}{LV-ViT-M}  \\
 \midrule
 Dense  & - & -  & 84.0  \\
 DynamicViT \cite{rao2021dynamicvit} & 0\% & 33.0\% & 83.8(-0.2) \\
 SPViT \cite{kong2021spvit} & 0\% &  42.2\%  &  83.7(-0.3)\\
\rowcolor{light-gray} Tri-Level   &  20.5\% &  24.1\%  & \textbf{83.8(-0.2)} \\
\bottomrule
\end{tabular}
}
\caption{Results on LV-VIT and PiT. We show the reduced training time, MACs (G) saving, and Top-1 accuracy comparison on the ImageNet-1K dataset.}
\label{lvvit_res}
\end{sc}
\end{small}
\end{center}
\end{table}

\begin{table}[htp]
\begin{center}
\begin{small}
\begin{sc}
\resizebox{0.9\columnwidth}{!}{
\begin{tabular}{l|ccc}
\toprule
\multirow{2}{*}{Backbone} & \multicolumn{3}{c}{Avg. Precision at IOU}    \\
     & AP & AP$_{50} $   &  AP$_{75}$ \\
\midrule
ResNet50 \cite{he2016deep} & 41.5 & 60.5 & 44.3 \\
ViT-S  & 36.9 & 57.0 & 38.0  \\
PiT-S  & 39.4 & 58.8 & 41.5 \\
 \rowcolor{light-gray} Tri-level  & 39.4 & 58.8 &  41.4 \\
\bottomrule
\end{tabular}
}
\caption{COCO detection performance based on Deformable DETR. We evaluate the performance of our method with ResNet50, ViT-S, and PiT-S as pretrained backbones for object detection.}
\label{tab:pit_coco}
\end{sc}
\end{small}
\end{center}
\end{table}




\end{document}